\documentclass{article} 
\usepackage{iclr2017_workshop,times}
\usepackage{hyperref}
\usepackage{url}
\usepackage{graphicx}
\usepackage{booktabs}       
\usepackage{amsfonts}       
\usepackage{nicefrac}       
\usepackage{microtype}      
\usepackage{subcaption}

\title{Fast Generation for Convolutional \\ Autoregressive Models}

\author{Prajit Ramachandran$^{1}$\thanks{Denotes equal contribution.}, Tom Le Paine$^{1}$\footnotemark[1], Pooya Khorrami$^1$, Mohammad Babaeizadeh$^1$, \\
{\bf Shiyu Chang$^2$, Yang Zhang$^1$,
 Mark A. Hasegawa-Johnson$^1$,} \\
{\bf Roy H. Campbell$^1$, \& Thomas S. Huang$^1$}\\
\\
$^1$University of Illinois at Urbana-Champaign, IL 61801, USA \\
\texttt{\small \{prmchnd2, paine1, pkhorra2, mb2, yzhan143,}\\
\texttt{jhasegaw, rhc, t-huang1\}@illinois.edu} \\
\\
$^2$IBM Thomas J. Watson Research Center, NY 10598, USA \\
\texttt{\small shiyu.chang@ibm.com}
}

\begin{document}

\maketitle

\begin{abstract}
Convolutional autoregressive models have recently demonstrated state-of-the-art performance on a number of generation tasks. While fast, parallel training methods have been crucial for their success, generation is typically implemented in a na\"{i}ve fashion where redundant computations are unnecessarily repeated. This results in slow generation, making such models infeasible for production environments. In this work, we describe a method to speed up generation in convolutional autoregressive models. The key idea is to cache hidden states to avoid redundant computation. We apply our fast generation method to the Wavenet and PixelCNN++ models and achieve up to $21\times$ and $183\times$ speedups respectively. 
\end{abstract}

\section{Introduction}


Autoregressive models are a powerful class of generative models that factorize the joint probability of a data sample $x$ into a product of conditional probabilities. Autoregressive models such as Wavenet \citep{oord2016wavenet}, ByteNet \citep{kalchbrenner2016neural}, PixelCNN \citep{oord2016pixel, oord2016conditional}, and Video Pixel Networks \citep{kalchbrenner2016video} have shown strong performance in audio, textual, image, and video generation. Unfortunately, generating in a na\"{i}ve fashion is typically too slow for practical use. For example, generating a batch of $16$ $32\times32$ images using PixelCNN++ \citep{salimans2017pixelcnn++} takes more then 11 minutes on commodity hardware with a Tesla~K40 GPU.

The ability to do fast generation is useful for many applications. Production environments have tight latency constraints, so real-time speech generation, machine translation, and image super-resolution \citep{dahl2017pixel} all require fast generation. Furthermore, quick simulation of environment dynamics is important for fast training in model-based reinforcement learning \citep{oh2015action}. However, slow generation hampers the use of convolutional autoregressive models in these situations. 

In this work, we present a method to significantly speed up generation in convolutional autoregressive models. The contributions of this work are as follows:
\begin{itemize}
\item[1.] We present a general method to enable fast generation for autoregressive models through caching. We describe specific implementations of this method for Wavenet~\citep{oord2016wavenet} and PixelCNN++~\citep{salimans2017pixelcnn++}. We demonstrate our fast generation achieves up to $21\times$ for Wavenet and $183\times$ for PixelCNN++ over their na\"{i}ve counterparts.
\item[2.] We open-source our implementation of fast generation for Wavenet\footnote{\url{https://github.com/tomlepaine/fast-wavenet}} and PixelCNN++\footnote{\url{https://github.com/PrajitR/fast-pixel-cnn}}. Our generation code is compatible with other open-source implementations of these models that also implement training. 
\end{itemize}

\section{Methods}

Na\"{i}ve generation for convolutional autoregressive models recalculates the entire receptive field at every iteration (we refer readers  to~\citet{oord2016wavenet,salimans2017pixelcnn++} for details). This results in exponential time and space complexity with respect to the receptive field. In this section, we propose a method that avoids this cost by caching previously computed hidden states and using them in the subsequent iterations.

We first start by describing how our method can speed up the generation of models with dilated convolutions. Then we discuss how to generalize our method to strided convolutions. Finally, we discuss the details of applying  our method to speed up Wavenet~\citep{oord2016wavenet} and  PixelCNN++~\citep{salimans2017pixelcnn++}.

\subsection{Caching for dilated convolutions}
\label{sec:cache:dilation}
\begin{figure}[ht]
  \centering
  \includegraphics[width=0.9\textwidth]{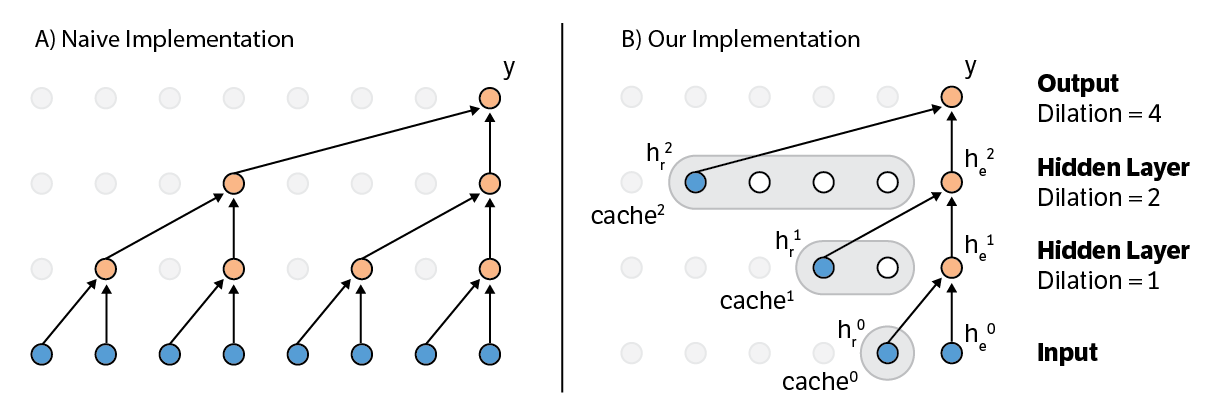}
  \caption{\textbf{Comparison of na\"{i}ve implementation of the generation process and our proposed method.} Orange nodes are computed in the current timestep, blue nodes are previously cached states, and gray nodes are not involved in the current timestep. Notice that generating a single sample requires $O(2^L)$ operations for the na\"{i}ve implementation where $L$ is number of layers in the network. Meanwhile, our implementation only requires $O(L)$ operations to generate a single sample.}
  \label{fig:naive_and_fast_gen}
\end{figure}

To generate a single output $y$, computations must be performed over the entire receptive field which is exponential with respect to the number of layers. A na\"{i}ve generation method repeats this computation over the entire receptive field at every step, which is illustrated in Figure \ref{fig:naive_and_fast_gen}A. However, this is wasteful because many hidden states in the receptive field can be re-used from previous iterations. This na\"{i}ve approach has been used in open-source implementations of Wavenet\footnote{\url{https://github.com/ibab/tensorflow-wavenet}}.

Instead of recomputing all of the hidden states for every iteration, we propose caching hidden states from previous iterations. Figure \ref{fig:naive_and_fast_gen}B illustrates this idea, where each layer maintains a cache of previously computed hidden states. During each generation step, hidden states are popped off the cache to perform the convolutions. Therefore, the computation and space complexity are linear in the number of layers instead of exponential.

\begin{figure}[h!]
\centering
\begin{minipage}{.47\textwidth}
  \centering
  \includegraphics[width=\linewidth, trim={0, 0, 0, 0}]{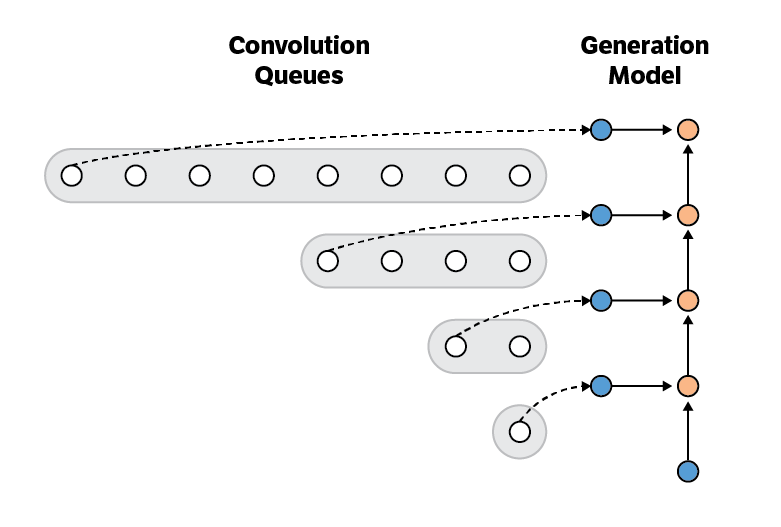}
  \captionof{figure}{\textbf{Pop phase for dilated convolutions.} The hidden states are popped off of each cache and fed as input (blue dots) into the corresponding location of the generation model. These hidden states and the current input (bottom blue dot) are used to compute the current output and the new hidden states (orange dots).}
  \label{fig:pop}
\end{minipage}%
\qquad 
\begin{minipage}{.47\textwidth}
  \centering
  \includegraphics[width=\linewidth, trim={0, 0, 0, 42}]{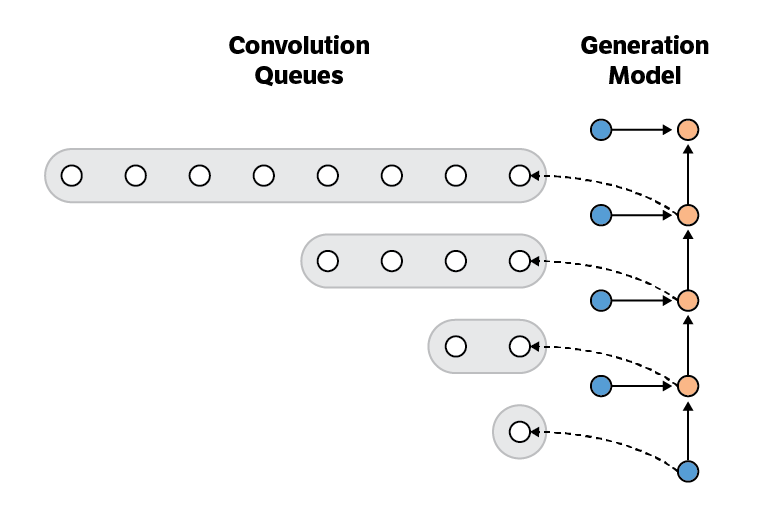}
  \captionof{figure}{\textbf{Push phase for dilated convolutions.} The new hidden states (orange dots) are pushed to the back of their respective caches. These hidden states will be used for future computations.}
  \label{fig:push}
\end{minipage}
\end{figure}

Figures \ref{fig:pop} and \ref{fig:push} demonstrate the caching mechanism in more detail. Each layer takes in its current input and a hidden state from the cache to compute the layer output. The cache is a queue in which the oldest hidden state is popped off to be fed into the current layer. The size of the cache is equivalent to the dilation of the layer. Thus, the oldest hidden state of the cache (i.e. front of the queue) is exactly one of the inputs that the layer should process. Finally, the output of the current layer is pushed into the cache (i.e. back of the queue) of the next layer to be used as input for future computation.

\subsection{Caching for strided convolutions}
\label{sec:strided_caching}
The caching algorithm for dilated convolutions is straightforward because  the number of hidden states in each layer is equal to the number of inputs. Thus, each layer can simply maintain a cache that is updated on every step. However, strided convolutions pose an additional challenge since the number of hidden states in each layer is different than the number of inputs.
 
\begin{figure}[ht]
  \centering
  \includegraphics[width=0.9\textwidth]{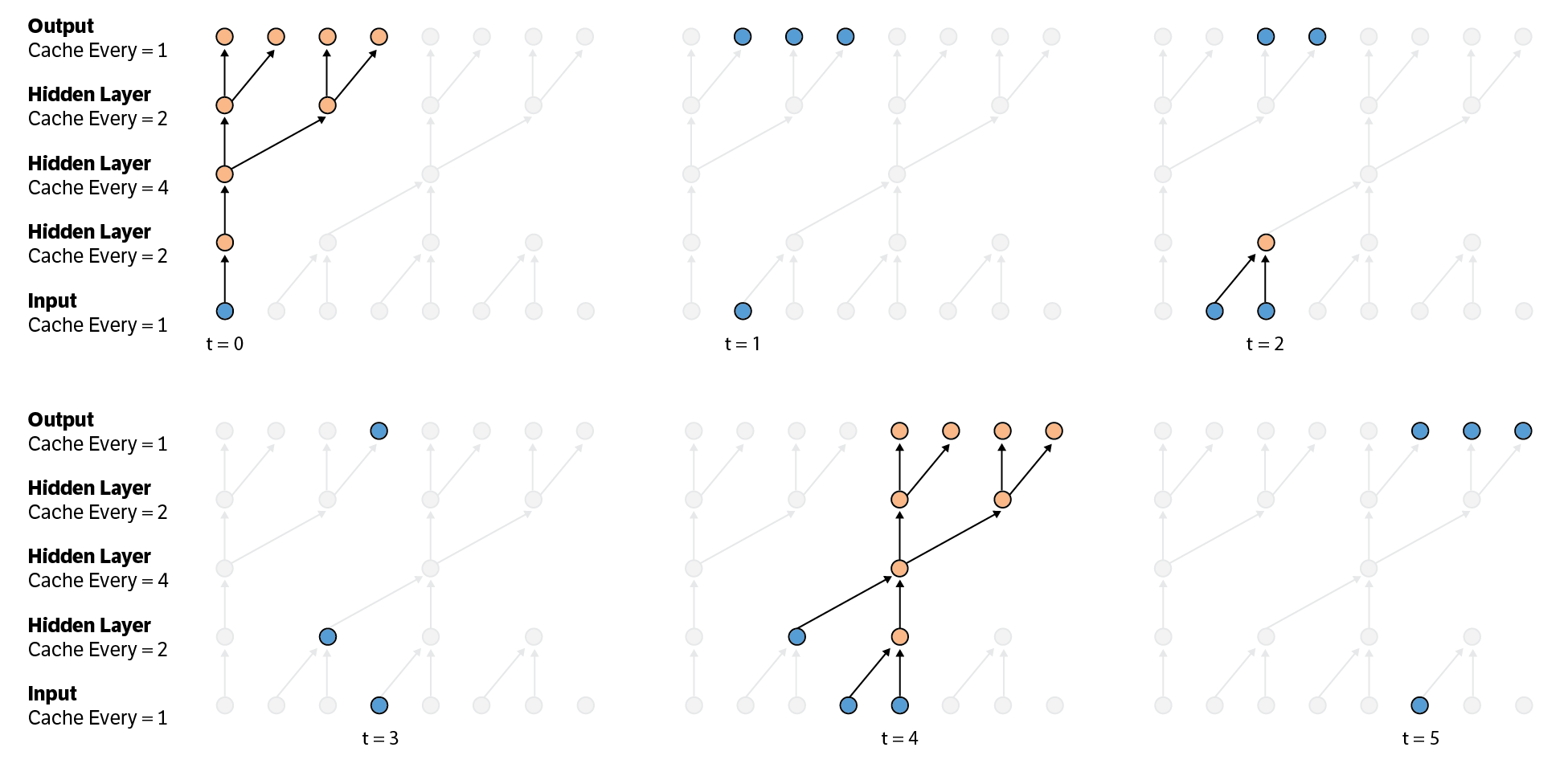}
  \caption{\textbf{Fast generation for a network with strided convolutions}. We show an example model with 2 convolutional and 2 transposed convolutional layers each with a stride of 2 \citep{dumoulin2016guide}. Due to the stride, each layer has fewer states than network inputs. Orange nodes are computed in the current timestep, blue nodes are previously cached states, and gray nodes are not involved in the current timestep. In the first timestep ($t = 0$), the first input is used to compute and cache all nodes for which there is sufficient information to generate, including the first four outputs. At $t = 1$, there are no nodes that have sufficient information to be computed, but the output for $t = 1$ has already been computed at $t = 0$. At $t = 2$, there is one new node that now has sufficient information to be computed, although the output for $t = 2$ has also been computed at $t = 0$. The $t = 3$ scenario is similar to $t = 1$. At $t = 4$, there is enough information to compute multiple hidden states and generate the next four outputs. This is analogous to the $t = 0$ scenario. $t = 5$ is analogous to $t = 1$, and this cycle is followed for all future time steps.}
  \label{fig:strided_conv}
\end{figure}

A downsampling (strided convolutional) layer will not necessarily generate  an output at each timestep (see the first hidden layer in Figure~\ref{fig:strided_conv}) and may even skip over some inputs (see the second hidden layer in Figure~\ref{fig:strided_conv}). On the other hand, an upsampling (strided transposed convolutional) layer will produce hidden states and outputs for multiple timesteps (see the last hidden layer in Figure~\ref{fig:strided_conv}). As a result, the cache cannot be updated in every timestep. Thus, each cache has an additional property \emph{cache\_every}, where the cache is only updated every \emph{cache\_every} steps. Every downsampling layer increases the \emph{cache\_every} property of the layer by the downsampling factor (2 in the case of Figure~\ref{fig:strided_conv}). Conversely, every upsampling layer decreases the \emph{cache\_every} property of the layer by the upsampling factor (also 2 in the case of Figure~\ref{fig:strided_conv}).

\subsection{Model-specific details}
Wavenet uses 1D dilated convolutions. Our fast implementation of Wavenet follows directly from the components outlined in Section~\ref{sec:cache:dilation}. 

PixelCNN++ improves upon PixelCNN~\citep{oord2016conditional} through a variety of modifications, including using strided convolutions and transposed convolutions instead of dilation for speed. Our method scales from 1D to 2D with very few changes. The caches for each layer are now 2D, with a height equal to the filter height and a width equal to the image width. After an entire row is generated, the oldest row of the cache is popped and the new row is pushed. Because strided convolutions are used, we use the \emph{cache\_every} idea detailed in Section~\ref{sec:strided_caching}. 

Furthermore, PixelCNN++ uses a vertical and horizontal stream (we refer readers to \citet{salimans2017pixelcnn++} for more details). Since the vertical stream does not depend on the horizontal stream, it is efficient to compute the vertical stream one entire row at a time and then use the cached vertical stream for every computation of the horizontal stream. For full details please refer to our code.

\section{Experiments}
We implemented our methods for Wavenet~\citep{oord2016wavenet} and PixelCNN++~\citep{salimans2017pixelcnn++} in TensorFlow~\citep{abadi2016tensorflow}. We compare our proposed method with a na\"{i}ve implementation of Wavenet\footnote{\url{https://github.com/ibab/tensorflow-wavenet}} and a na\"{i}ve implementation of PixelCNN++\footnote{\url{https://github.com/openai/pixel-cnn}}. In the case of PixelCNN++, we vary the number of images generated in parallel (\emph{batch size}), mirroring batching in production environments. As the batch size increases, our method significantly outperforms the na\"{i}ve implementation in runtime. For example, on batch sizes of 128 and 256, our method is two orders of magnitude faster. In the Wavenet experiment, the batch size has been fixed to 1 in order to highlight the effect of adding layers on performance.  The results indicate significant speedups, up to $21\times$ for Wavenet and $183\times$ for PixelCNN++.

\begin{figure}[h!]
\centering
\begin{minipage}{.45\textwidth}
  \centering
  \includegraphics[width=\linewidth, trim={0, 0, 0, 0}]{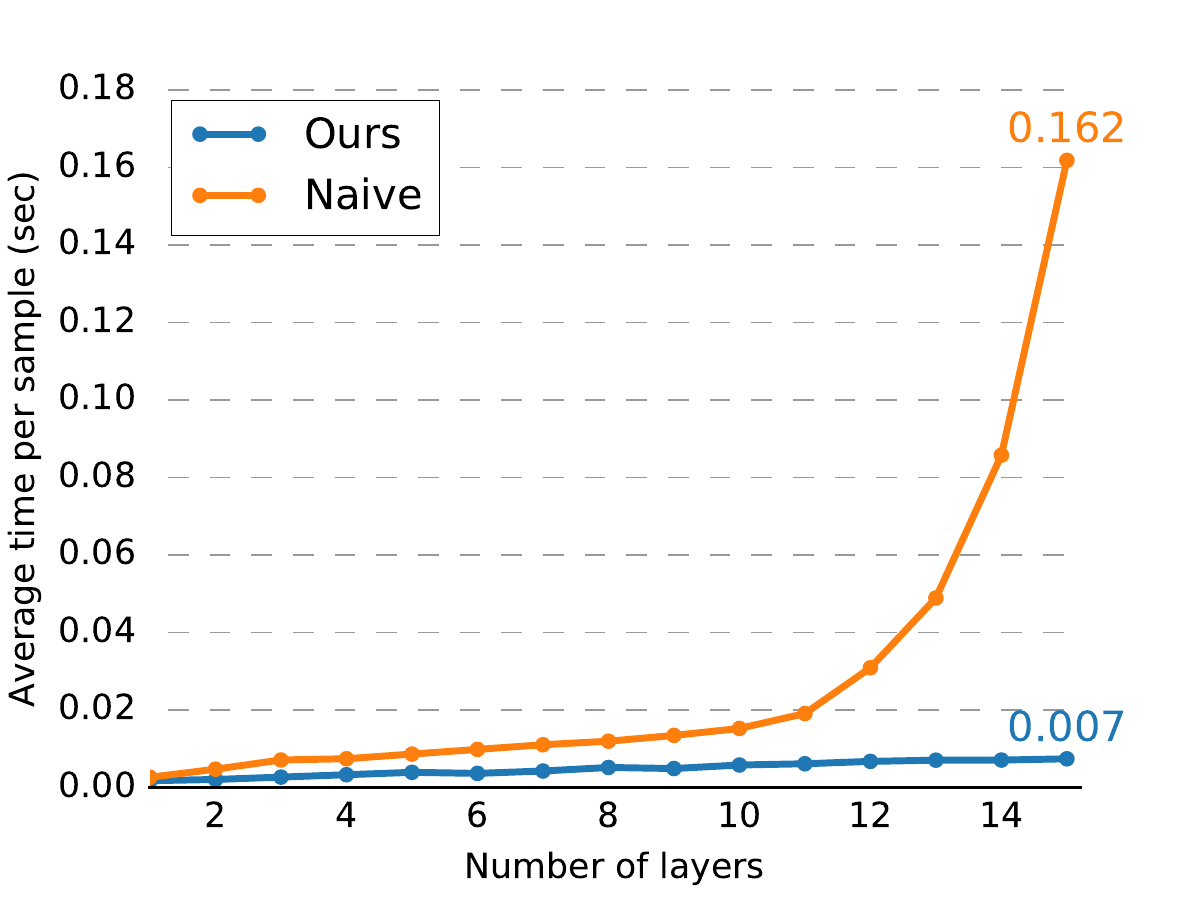}
  \captionof{figure}{\textbf{Wavenet timing experiments.} We generated from a model with $2$ sets of $L$ dilation layers each, using a na\"{i}ve implementation and ours. Results are averaged over $100$ repeats. When $L$ is small, the na\"{i}ve implementation performs better than expected due to GPU parallelization of the convolution operations. When $L$ is large, the difference in performance is more pronounced.
}
  \label{fig:timing_experiments_wavenet}
  \label{fig:wavenet_speedup}
\end{minipage}%
\qquad 
\begin{minipage}{.49\textwidth}
  \centering
  \includegraphics[width=\linewidth, trim={0, 0, 0, 40}]{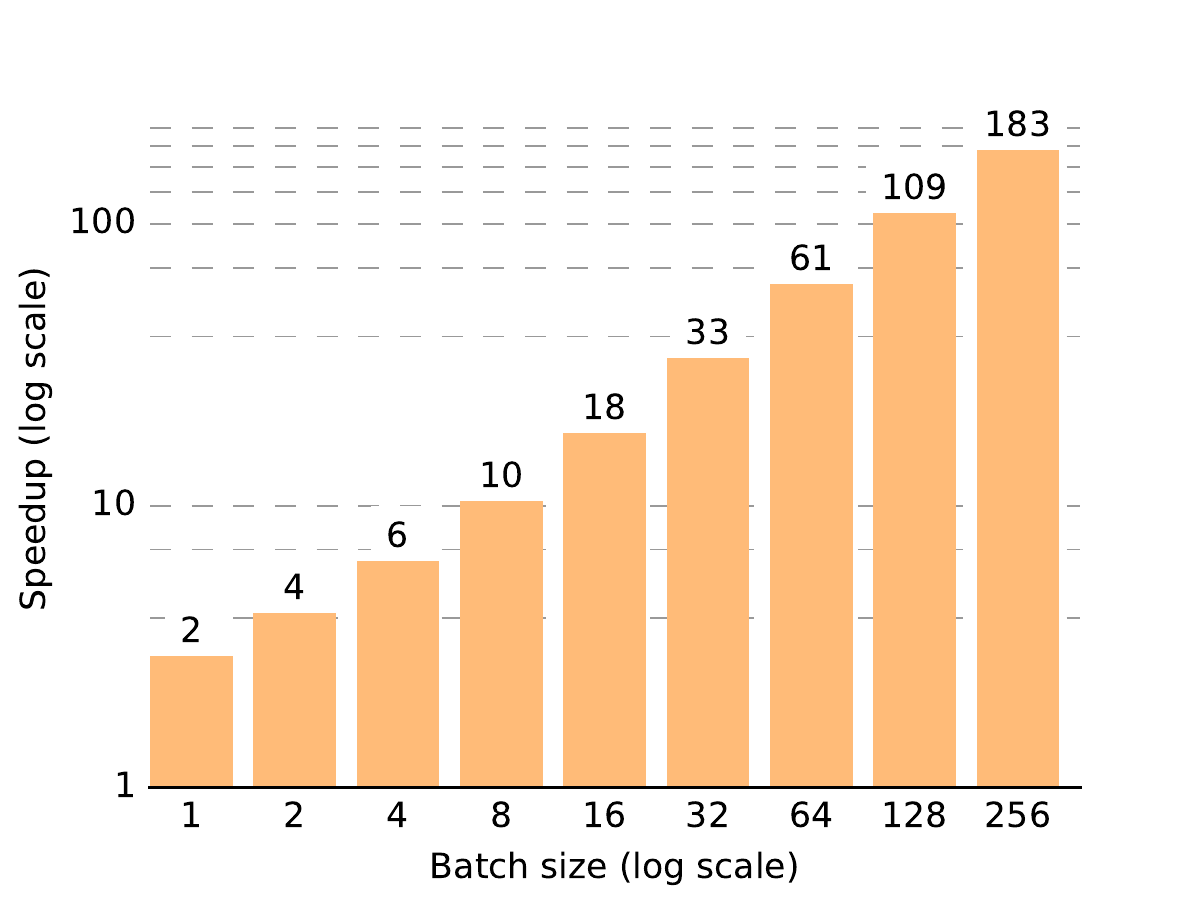}
  \captionof{figure}{\textbf{PixelCNN++ timing experiments.} We generated images using the model architecture described in~\citep{salimans2017pixelcnn++}. Due to the huge number of convolution operations in the na\"{i}ve implementation, GPU utilization is always high and there is no room for parallelization across batch. Since our method avoids redundant computations, larger batch sizes result in larger speedups. 
  }
  \label{fig:pixelcnn_speedup}
\end{minipage}
\end{figure}



\subsubsection*{Acknowledgments}
Authors would like to thank Wei Han and Yuchen Fan for their insightful discussions, as well as Maize group for their support. The Tesla K40 GPU used for this research was donated by the NVIDIA Corporation.


\bibliographystyle{iclr2017_workshop}
\bibliography{iclr2017_workshop}


\end{document}